\documentclass[letterpaper, 10 pt, journal, twoside]{IEEEtran}
\usepackage{cite}
\usepackage{amsmath,amssymb}
\usepackage{graphicx}
\usepackage{textcomp}
\usepackage{xcolor}
\usepackage[ruled,vlined]{algorithm2e}
\usepackage{hyperref}
\DeclareMathOperator*{\argmin}{arg\,min}
\usepackage{cleveref}
\crefname{section}{§}{§§}
\Crefname{section}{§}{§§}
\usepackage{fancyhdr}

\begin{document}

\title{Learning of Causal Observable Functions for Koopman-DFL Lifting Linearization of Nonlinear Controlled Systems and Its Application to Excavation Automation}

% \author{Michael~Shell,~\IEEEmembership{Member,~IEEE,}
%         John~Doe,~\IEEEmembership{Fellow,~OSA,}
%         and~Jane~Doe,~\IEEEmembership{Life~Fellow,~IEEE}% <-this % stops a space
% \thanks{M. Shell was with the Department
% of Electrical and Computer Engineering, Georgia Institute of Technology, Atlanta,
% GA, 30332 USA e-mail: (see http://www.michaelshell.org/contact.html).}% <-this % stops a space
% \thanks{J. Doe and J. Doe are with Anonymous University.}% <-this % stops a space
% \thanks{Manuscript received April 19, 2005; revised August 26, 2015.}}
\author{Nicholas~Stearns~Selby and H.~Harry~Asada,~\IEEEmembership{Senior~Member,~IEEE,}~%
\thanks{Manuscript received: March 1, 2021; Revised May 20, 2021; Accepted June 14, 2021. This paper was recommended for publication by Editor Youngjin Choi upon evaluation of the Associate Editor and Reviewers' comments.
This material is based upon work supported by National Science Foundation Grant NSF-CMMI 2021625. The authors are with the School of Engineering, Massachusetts Institute of Technology, Cambridge, MA, USA
        {\tt\footnotesize nselby@mit.edu; asada@mit.edu}}%
\thanks{Digital Object Identifier (DOI): 10.1109/LRA.2021.3092256.}
\thanks{© 2021 IEEE.  Personal use of this material is permitted.  Permission from IEEE must be obtained for all other uses, in any current or future media, including reprinting/republishing this material for advertising or promotional purposes, creating new collective works, for resale or redistribution to servers or lists, or reuse of any copyrighted component of this work in other works.}
}

\markboth{IEEE Robotics and Automation Letters. Preprint Version. Accepted June 2021}
{Selby and Asada: Learned Lifting Linearization} 




\maketitle

\begin{abstract}
Effective and causal observable functions for low-order lifting linearization of nonlinear controlled systems are learned from data by using neural networks. While Koopman operator theory allows us to represent a nonlinear system as a linear system in an infinite-dimensional space of observables, exact linearization is guaranteed only for autonomous systems with no input, and finding effective observable functions for approximation with a low-order linear system remains an open question. Dual-Faceted Linearization uses a set of effective observables for low-order lifting linearization, but the method requires knowledge of the physical structure of the nonlinear system. Here, a data-driven method is presented for generating a set of nonlinear observable functions that can accurately approximate a nonlinear control system to a low-order linear control system. A caveat in using data of measured variables as observables is that the measured variables may contain input to the system, which incurs a causality contradiction when lifting the system, i.e. taking derivatives of the observables. The current work presents a method for eliminating such anti-causal components of the observables and lifting the system using only causal observables. The method is applied to excavation automation, a complex nonlinear dynamical system, to obtain a low-order lifted linear model for control design.
\end{abstract}

% \begin{IEEEkeywords}
% IEEE, IEEEtran, journal, \LaTeX, paper, template.
% \end{IEEEkeywords}

% \fi
\IEEEpeerreviewmaketitle

\section{Introduction}
\label{sec:intro}
%Nonlinearity is everywhere in robotics. Ranging from nonlinear dynamics of robotic arms and legs to mechanical contact with friction and contact-noncontact discontinuity, governing equations of robots and their environment conditions are nonlinear, posing unique challenges in broad robotics problems. Many of these complex nonlinearities are difficult to parameterize. Data-driven methods are therefore increasingly important, as robotics are applied to complex processes. 

\IEEEPARstart{T}{here} is a growing need in the construction and mining industries for excavation automation. Various technologies are being developed for operating excavators autonomously with increased productivity and fuel
efficiency \cite{Dadhich-etal}. The recent and projected growth of the global construction industry \cite{economics2015global} and the dangers of the excavation work environment \cite{sotiropoulos2019dig-data} are major drivers behind the development of intelligent excavators for performing earth-moving tasks.

\begin{figure}
    \centering
    \includegraphics[width=\linewidth]{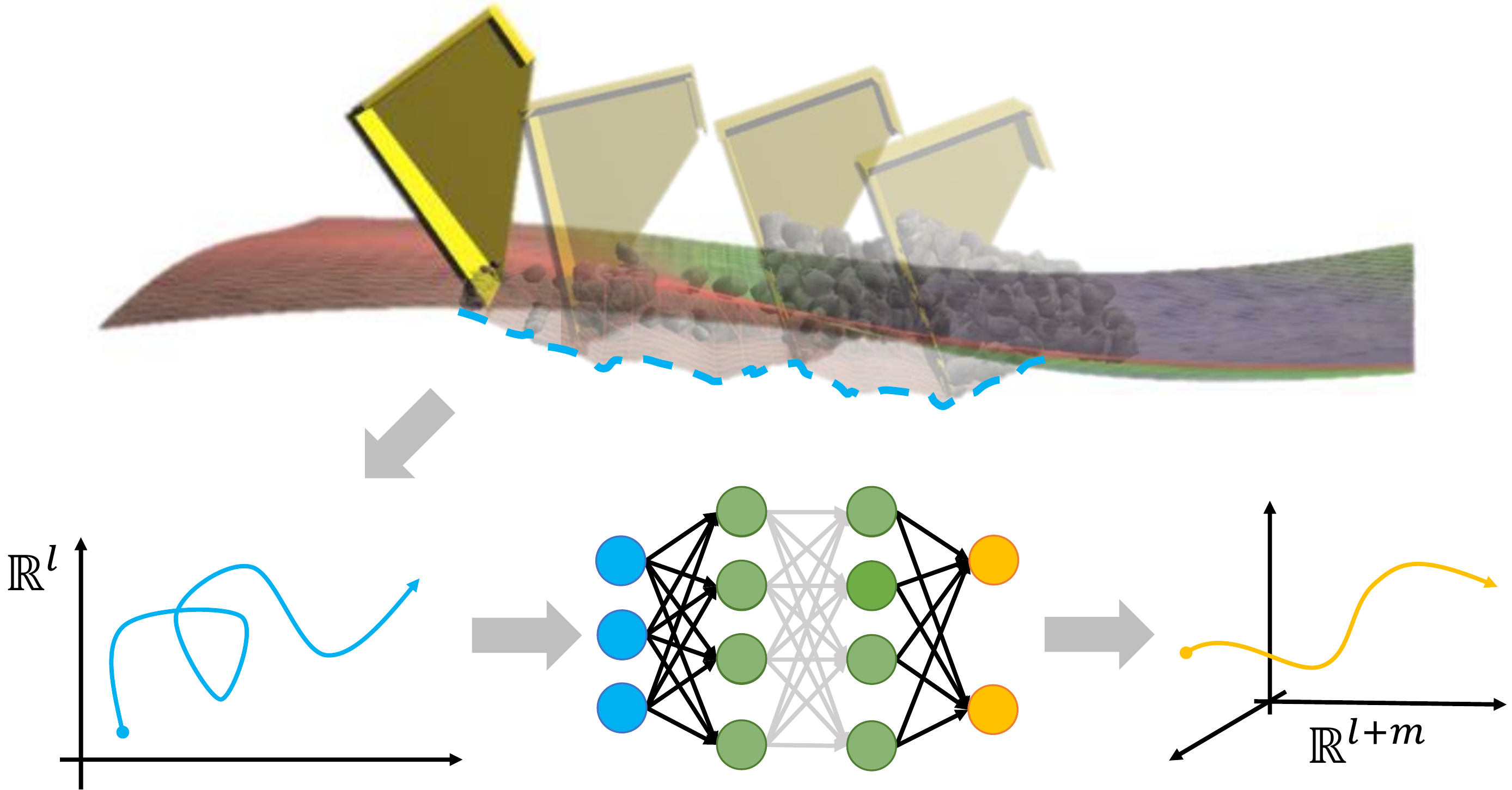}
    \caption{Learned lifting linearization of autonomous excavation.}
    \label{fig:agx}
\end{figure}

Excavation is a highly nonlinear process where soil and rocks interact with the bucket of an excavator in a complex manner (see Fig. \ref{fig:agx}). While terramechanics models have been studied for many decades, their validity is limited due to the difficulty of identifying the numerous parameters of mechanistic models. Data-driven methods have recently been introduced to autonomous excavation for capturing complex nonlinearities \cite{dig-data1,dig-data2,sandzimier2020dig-data,sotiropoulos2019dig-data,jud2019autonomous}, yet the nonlinear models are still too complex to use, in particular, for real-time control.

Lifting linearization is a methodology for representing a nonlinear dynamical system with a linear dynamic model in a high-dimensional space. Underpinned by Koopman operator theory, nonlinear systems represented with supernumerary state variables behave more linearly in the lifted space. The method has recently been applied to various robotics and automation challenges, including active learning \cite{abraham2019active}, soft robotics \cite{soft-robot-koop}, human-robot interaction \cite{learn-koop-hri}, power systems \cite{koopman-app1}, and mission planning \cite{koopman-app3}. More broadly, deep learning has proven a valuable tool for lifting linearization techniques \cite{zhang2019solar,ichter2019robot,haarnoja2016backprop,watter2015embed}.

The original Koopman Operator has two major limitations:
\begin{enumerate}
    \item The theory is applicable only to dynamical systems with no exogenous input, i.e. autonomous systems, and
    \item Exact linearization requires an infinite-dimensional space, except for a restricted class of systems.
\end{enumerate}
Any extension to non-autonomous, finite-dimensional systems is no longer exact, but an approximation. Various methods for truncating the system with a finite-dimensional space have been reported. Among others, the eigendecomposition of the lifted system allows us to represent the system at desirable accuracy and granularity while providing useful insights into the system \cite{koopman-book}. Furthermore, the extended Dynamic Mode Decomposition (eDMD) is completely data-driven, providing a practical tool for complex nonlinear system representation \cite{vijayshankar2020dynamic}. These methods, however, need a set of observables, i.e. output variables, which are nonlinear functions of independent state variables. It is still an open question how to find an effective set of observable functions.

One of the key challenges in the lifting linearization of nonlinear systems with exogenous input is causality. If observable functions are functions of both state variables and input variables, we cannot use such observables for lifting the system. Lifting entails computing time derivatives of the observables and, thereby, the dynamic equations inevitably include the time derivative of input. In discrete formulation, this means the use of future input. Including these input terms, the time-evolution of the observables turns out not to be causal. If one measures a set of candidates of observables from a nonlinear system that is subject to control inputs and uses the measured variables for lifting the system, they may end up with a non-causal dynamical system.

In Dual-Faceted Linearization (DFL), another approach to lifting linearization, the causality issue is analyzed based on physical system modeling theory \cite{asada-dfl}. In DFL, the propagation of inputs across the nonlinear dynamical system can be tracked, and their effect on all observables, called auxiliary variables, can be localized. Assuming that inputs are linearly involved in observables, a method has been established for eliminating the input-dependent component from each observable and lifting the dynamics using the remaining input-free observables. In the Koopman-based lifting linearization, too, it is assumed that the observables are input-affine in order to eliminate input-dependent components from observable functions so a causal dynamic model can be obtained \cite{korda2018linear}.

Lifting linearization is a powerful methodology for tackling a broad spectrum of nonlinear problems, in particular, excavation process modeling and control. However, two critical challenges have not yet been fully solved:
\begin{enumerate}
    \item Finding an effective set of observables to approximate a nonlinear system in a low-dimensional lifted space
    \item Finding causal observables uncorrelated with inputs
\end{enumerate}
The objective of the current work is to solve these two challenges. We present a low-dimensional, causal, lifting linear model obtained from experimental data. Neural networks are used to find effective observables through learning.

In the following, we summarize a basic formulation of lifting linearization in \cref{sec:bkgnd}. We present the learning method for obtaining an effective set of observables in \cref{sec:algo}. First, we deal with nonlinear controlled systems where all measured observables are not affected by inputs. Then, the method is extended for physical observables that may be functions of inputs. Simple numerical examples are discussed for validating the proposed method in \cref{sec:toy}, and we apply it to excavation process modeling in \cref{sec:terra}.
\section{Background}
\label{sec:bkgnd}
This section summarizes background knowledge for readability. More details can be found in \cite{koopman1931hamiltonian,asada-dfl}.

\subsection{Koopman Operator Theory}
First proposed in 1931 by Koopman \cite{koopman1931hamiltonian}, Koopman operator theory originally modeled autonomous systems by mapping nonlinear dynamics onto an infinite-dimensional linear space. Later techniques expanded the use of the operator for nonautonomous systems \cite{koopman-nonautonomous} and developed methods for approximating the infinite-dimensional mapping with a computationally feasible, finite-dimensional space \cite{koop-finite1,koop-finite2,koop-finite3}.

Let the discrete-time dynamics of a nonlinear, autonomous system with state $x_t\in \mathbb{R}^l$ at time $t$ be given by $x_{t+1}=f(x_t)$. Furthermore, define a vector of nonlinear observables of the state, $\eta_t=g(x_t) \in \mathbb{R}^m$.

The Koopman operator, $\mathcal{K}$, is linear and infinite-dimensional and applies to observable functions: $\mathcal{K}_fg=g\circ f$, where $\circ$ represents the composition operator.

\subsection{Dual-Faceted Linearization (DFL)}
Despite the use of Koopman operators to provide a lifting linearization for autonomous systems, the theory provides no method by which to select an effective set of observables. Because it is infeasible to compute the infinite-dimensional space with finite computational resources, the choice of which observables to use is very important. DFL \cite{asada-dfl} uses a particular class of observables that are determined based on physical modeling theory and bond graphs \cite{bond-graph-textbook}. Those observables, called auxiliary variables, are physically meaningful, and may be measured physically. Furthermore, causality analysis of the method allows us to examine how exogenous inputs propagate the system and influence specific auxiliary variables. Using those variables with no input influence, one can obtain a lifted system that is causal. Alternatively, input-dependent variables can be ``laundered'' into causal variables.

Consider the discrete-time dynamics of a nonlinear, nonautonomous system with input $u_t\in \mathbb{R}^n$ at time $t$ given by:
\begin{equation}
    x_{t+1}=f(x_t,u_t)
    \label{eq:disc-dyn-aut}
\end{equation}

Assuming that the system is a lumped-parameter system with integral causality,  we can choose outputs of all the nonlinear elements involved in the system as observables $\eta_t=g(x_t)\in \mathbb{R}^m$ to augment the system state and construct a linear representation of the system dynamics:
\begin{equation}
    x_{t+1}=A_x x_t+A_\eta \eta_t+B_x u_t
    \label{eq:dfl-x-dyn}
\end{equation}
where $A_x\in \mathbb{R}^{l\times l}$, $A_\eta\in \mathbb{R}^{l\times m}$, and $B_x\in \mathbb{R}^{l\times n}$ are fixed matrix coefficients determined by the physical structure of the system. This part of the state evolution is exact.

We approximate the $\eta$-dynamics using a second equation:
\begin{equation}
    \eta_{t+1}=H_x x_t+H_\eta \eta_t+H_u u_t+r_{\eta_{t+1}}
    \label{eq:dfl-eta-dyn}
\end{equation}
where $H_x\in \mathbb{R}^{m\times l}$, $H_\eta\in \mathbb{R}^{m\times m}$, and $H_u\in \mathbb{R}^{m\times n}$ are fixed matrix coefficients and $r_{\eta_{t+1}}\in \mathbb{R}^m$ is a residual. Unlike $A_x$, $A_\eta$, and $B_x$, which are determined from the physical structure of the system, $H_x$, $H_\eta$, and $H_u$ must be regressed from data. For brevity, define coefficient matrices $A\triangleq \left( A_x, A_\eta, B_x \right)\in \mathbb{R}^{l\times p}$ and $H\triangleq \left( H_x, H_\eta, H_u \right)\in \mathbb{R}^{m\times p}$ and datum vector $\xi_t\triangleq \left( x_t^\intercal, \eta_t^\intercal, u_t^\intercal \right)^\intercal \in \mathbb{R}^p$ where $p=l+m+n$. Apply a negative discrete-time shift operator $T_{-1}$ to (\ref{eq:dfl-eta-dyn}) to optimize $H$ to minimize the mean squared error of predicting $\eta_{t+1}$:
\begin{equation}
    \begin{array}{rl}
        H^\mathrm{o} & =\argmin_H \mathrm{E}\left[ \left| H\xi_{t-1}-\eta_t \right|^2 \right] \\
         & =\mathrm{E}\left[ \eta_t \xi_{t-1}^\intercal \right]\left( \mathrm{E}\left[ \xi_{t-1} \xi_{t-1}^\intercal \right] \right)^{-1}
    \end{array}
    \label{eq:dfl-h-opt}
\end{equation}
where $\mathrm{E}[\cdot ]$ is the expectation operator. Assuming that the system is persistently excited and that $u_t$ is not collinear with $x_t$, there is a unique solution, $H^\mathrm{o}$.

\begin{figure*}
    \centering
    \includegraphics[width=\linewidth]{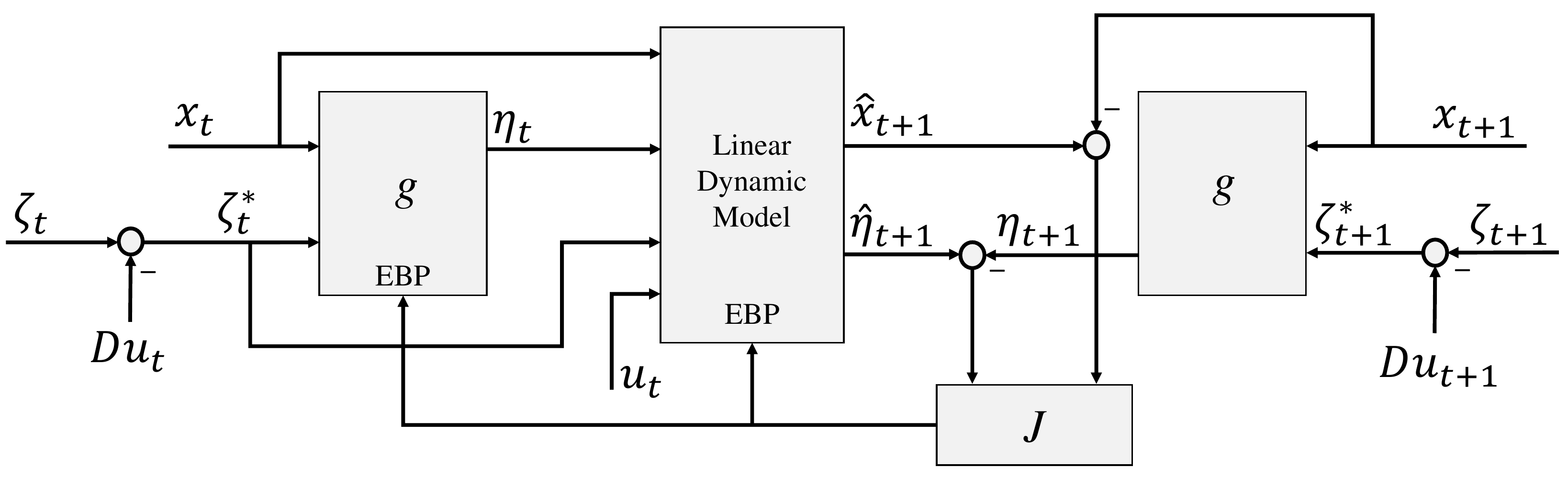}
    \caption{Block diagram of the learned lifting linearization algorithm. The loss, $J$, is used to tune the weights of neural network, $g$, and linear dynamic model matrices $A$ and $H$ via error backpropagation (EBP). Note that both instances of $g$ are equivalent to each other for all time $t$.}
    \label{fig:blocks}
\end{figure*}

The original nonlinear dynamics $f$ can now be modeled using the dual-faceted linear dynamics:
\begin{equation}
    x_{t+1}= A\xi_t; \quad \eta_{t+1}\approx H\xi_t; \quad \eta_0=g(x_0);
    \label{eq:cases-dfl}
\end{equation}

Practical benefits of using DFL as a tool to model systems include:
\begin{itemize}
    \item Augmented state feedback can be used to better inform controllers \cite{jerry-dfl}.
    \item Linear observer design is enabled for augmented state feedback.
    \item Model-predictive control is convex \cite{koopman-book}.
    \item Because DFL is based in physical modelling theory, augmented state systems may be measurable, and dynamics may have physical intuition \cite{filippos-dfl}.
\end{itemize}
For these reasons, DFL has proven to be a valuable tool in modeling nonlinear systems. However, DFL requires knowledge of the structure of the physical system. Furthermore, many systems contain no obvious, measurable observables with which to augment the state, and there is no guarantee that, when they do exist, physically meaningful observables make the best choices for augmenting the system state.

\subsection{Machine Learning for Linear Latent Spaces}
Like Koopman operator theory and DFL, learned latent-space dynamic modeling techniques also involve constructing a nonlinear representation of the original state, then using a model to evolve the new ``observables'' through time. In Koopman operator theory and DFL, the dynamic model is linear. Recently, much work has been done to explore applying deep learning techniques to Koopman operators.

Abraham and Murphey \cite{abraham2019active} presented an active learning strategy for robotic systems that extended observables, $g$, from Koopman operator theory to include the control input, $u_t$. Their algorithm trains a neural network to approximate an optimal $g$. In each epoch, they recompute a finite-dimensional matrix approximation of the Koopman operator $\mathcal{K}$, but because $g$ is a function of both $x_t$ and $u_t$, regressing such a matrix requires knowing $u_{t+1}$ in addition to $x_{t+1}$. To solve this causality problem, they propose replacing $u_{t+1}$ with $u_t$ in the Koopman operator update.

Han \textit{et al.} \cite{learn-koop} also proposed using a neural network that approximates an optimal lifting function $g$. At the end of each epoch, after feeding $x_{t-1}$ forward through $g$, they solve a least squares optimization problem to regress a modified $H^\mathrm{o}$ where $H_x=0$, then use the learned model to backpropagate the error $\mathrm{E}\left[ \left| \left| g(x_t)-H^\mathrm{o}\xi_{t-1} \right| \right|_F \right]$ plus a penalty on the norms of the components of $H^\mathrm{o}$. Because they do not track the evolution of the state, $x$, directly, they must simultaneously learn an additional matrix to approximate $g^{-1}$.

Lusch \textit{et al.} \cite{autoencoder1} and Mastia and Bemporad \cite{autoencoder2} replaced the $g^{-1}$ matrix approximation of Han \textit{et al.} with a neural network decoder, while in this work we handle this by feeding the state forward directly into the linear dynamic model.
Their models learn to minimize cost along three axes: neural network $g^{-1}$ must invert neural network $g$, i.e. $x=g\left(g^{-1}(x)\right)$; the nonlinear model must be able to predict, i.e. $x_{t+1}=g^{-1}\left(Hg(x_t)\right)$; and the lifted state must propagate linearly, i.e. $g(x_{t+1})=Hg(x_t)$.

Work by Yeung \textit{et al.} \cite{learned-dmd} applied deep learning to dynamic mode decomposition (DMD) by training a neural network to approximate the ``snapshot'' function mapping state to physical observable. Instead of optimizing a complicated cost function like in \cite{learn-koop, autoencoder1, autoencoder2}, their work simply trains a model to map states to other measurements of the system, then regresses a linear dynamic model to propagate the measurements forward in time.

Other work in learned latent spaces for lifting linearizations leverages neural networks to approximate functions similar to Koopman's observables in reinforcement learning \cite{watter2015embed}, sampling-based motion planning \cite{ichter2019robot}, Kalman filtering \cite{haarnoja2016backprop}, and partially observable Markov decision process \cite{zhang2019solar} paradigms. However, these works all learn linearizations $A$ and $H$ that are time- or state-dependent, further reducing the long-term robustness of the linear model and giving up the benefits of provably convex optimal control.

\subsection{Anticausal Observables}
Most lifting linearization techniques, including Koopman operator theory and DFL, require that the lifting observables be control input-independent, i.e. $\eta=g(x)$. If auxiliary variables depend on the control input, $u$, i.e. $\eta=g(x,u)$, then propagating $\eta$ forward through time requires knowledge of future values of control input:
\begin{equation}
    \eta_{t+1}\approx \eta_t+\frac{\partial \eta}{\partial x} \left( x_{t+1}-x_t \right)+\frac{\partial \eta}{\partial u}\left( u_{t+1}-u_t\right)
\end{equation}

In order to avoid problems with causality, most methods explicitly avoid augmenting the system state with control input-dependent variables. There are two common techniques.

The first solution to the causality problem is to include a state-feedback control law in the model \cite{abraham2019active,learn-koop}. By constraining the control input to be a known function of state, the system becomes effectively autonomous, and the original formulation of Koopman operator theory applies. This works for regulators, including for controllers regulating a system to follow a predetermined state trajectory, but no exogenous input is allowed.

The second solution to the causality problem is to assume that the auxiliary variables are linear in $u$ \cite{korda2018linear,asada-dfl,mamakoukas2019local}:
\begin{equation}
    \eta(x,u) = \eta^*(x)+Du
    \label{eq:eta-du}
\end{equation}
where $\eta^*$ is exclusively state-dependent and $D$ is a fixed matrix coefficient of $u$.

Although predicting future values of $\eta(x,u)$ remains impossible without a control law, this formulation allows for the modeling of the evolution of $\eta^*(x)$. The auxiliary state equation can be rewritten:
\begin{equation}
    \eta^*_{t+1}=H_x^* x_t+H_\eta^* \eta_t+H_u^* u_t+r_{\eta^*,t}
    \label{eq:etas-dyn}
\end{equation}
where $r_{\eta^*,t}$ is a residual.

Substituting (\ref{eq:eta-du}) into (\ref{eq:etas-dyn}) yields:
\begin{equation}
    \eta^*_{t+1}=H_x^* x_t+H_\eta^* \eta^*_t+\left( H_u^*+H_\eta^*D \right) u_t+r_{\eta^*,t}
\end{equation}
which is a causal, augmented state dynamic equation. The question of causality is therefore solved by preprocessing the auxiliary state data to filter out their dependence on $u$.
% \begin{figure*}
%     \centering
%     \includegraphics[width=\linewidth]{figures/blocks.pdf}
%     \caption{Block diagram of the learned lifting linearization algorithm. The loss, $J$, is used to tune the weights of neural network, $g$, and linear dynamic model matrices $A$ and $H$ via error backpropagation (EBP). Note that both instances of $g$ are equivalent to each other for all time $t$.}
%     \label{fig:blocks}
% \end{figure*}

\section{Modeling Algorithm}
\label{sec:algo}
% In this paper, we seek to apply the modeling benefits of DFL to learned latent spaces. Note that the finite-dimensional matrix representation of the Koopman operator $\mathcal{K}$ is related to the coefficient matrix $H$ in DFL. Because Koopman operators were originally only applied to autonomous systems, the domain of $\mathcal{K}_fg$ is $\mathbb{R}^m$. Because DFL models nonautonomous systems and uses $\eta$ to augment, rather than replace, the state, the domain of $H$ is larger, $\mathbb{R}^{l+m+n}$, enabling a higher-dimensional linear parametrization of the dynamics. By treating all parameters of the DFL model, $g$, $A$, and $H$, as learnable, we can avoid the modeling pitfalls of previous works, enabling a data-driven, time- and state-independent, lifted linearization algorithm capable of directly tracking the system state without dictating a control algorithm.

% Because we treat $A$ as a learnable parameter, in addition to minimizing the residual term, $r_\eta(x_t,u_t)$, we must override the state transition function from DFL in Eq. \ref{eq:dfl-x-dyn} to include a residual term as well:
% \begin{equation}
%     x_{t+1}=A_x x_t+A_\eta \eta_t+B_x u_t+r_x(x_t,u_t)
%     \label{eq:dfl-x-dyn-res}
% \end{equation}

% By redefining the goal of the modeling problem purely to minimize the dual residual of the augmented linear model, data-driven state augmentation should produce more exact results than traditional DFL and previous work on machine learning for linear latent spaces.

% \subsection{Learning Framework}

Fig. \ref{fig:blocks} shows the overview of the modeling algorithm. The learning system consists of three major components. The first is the Linear Dynamic Model predicting the transition of the system. The second is a multi-layer neural network for generating auxiliary variables, and the third is the process of evaluating the prediction error.

We assume all state variables are accessible and sufficient data for training are attainable. All state variables are fed into the left neural network, $g$, to produce a set of synthetic observables, $\eta_t$, to be learned. The state $x_t$, observables $\eta_t$, and input $u_t$ are fed into the linear dynamic model parameterized by matrices $A$ and $H$. The linear model produces predicted state $\hat{x}_{t+1}$ and predicted observables $\hat{\eta}_{t+1}$. The predicted state and observables are compared to their ground truth values, $x_{t+1}$ and $\eta_{t+1}=g(x_{t+1};\theta)$, respectively. The right neural network in the figure is a twin copy of the left neural network, sharing parameters $\theta$.

The squared error of the predicted state and observables, $J$, is used for updating the linear dynamic model with respect to parameters $A$ and $H$ and the neural network weights. The update of these parameters is computed via error backpropagation (EBP).

The causality analysis involved in the DFL modeling allows us to examine whether  observables, called auxiliary variables, are functions of state alone or include inputs. If some auxiliary variables are causal, having no dependence on control input, they can be added to the state for lifting the dynamics. Let $\zeta^*_t$ represent causal auxiliary variables. As shown in Fig. \ref{fig:blocks}, the causal observables can be fed into the neural network, so that the synthetic auxiliary variables $\eta_t$ can be produced from richer data. Note that the ground truth $\eta_{t+1}$, too, can be produced in response to not only state $x_{t+1}$ but also $\zeta_{t+1}^*$. The tunable parameter space of the neural network is expanded with the use of the causal auxiliary variables.

In case physically measurable auxiliary variables are functions of both state and input, such input-dependent auxiliary variables cannot be used in their original form for lifting the dynamics. It is necessary to filter out the input components from the observables. Fig. \ref{fig:blocks} also shows a simple filter to eliminate the effect of input from those variables, $\zeta_t$, as discussed in \cref{sec:anticausal}.

\subsection{Discrete-Time Learned Lifting Linearization}
\begin{figure}[b!]
    \centering
    \includegraphics[width=\linewidth]{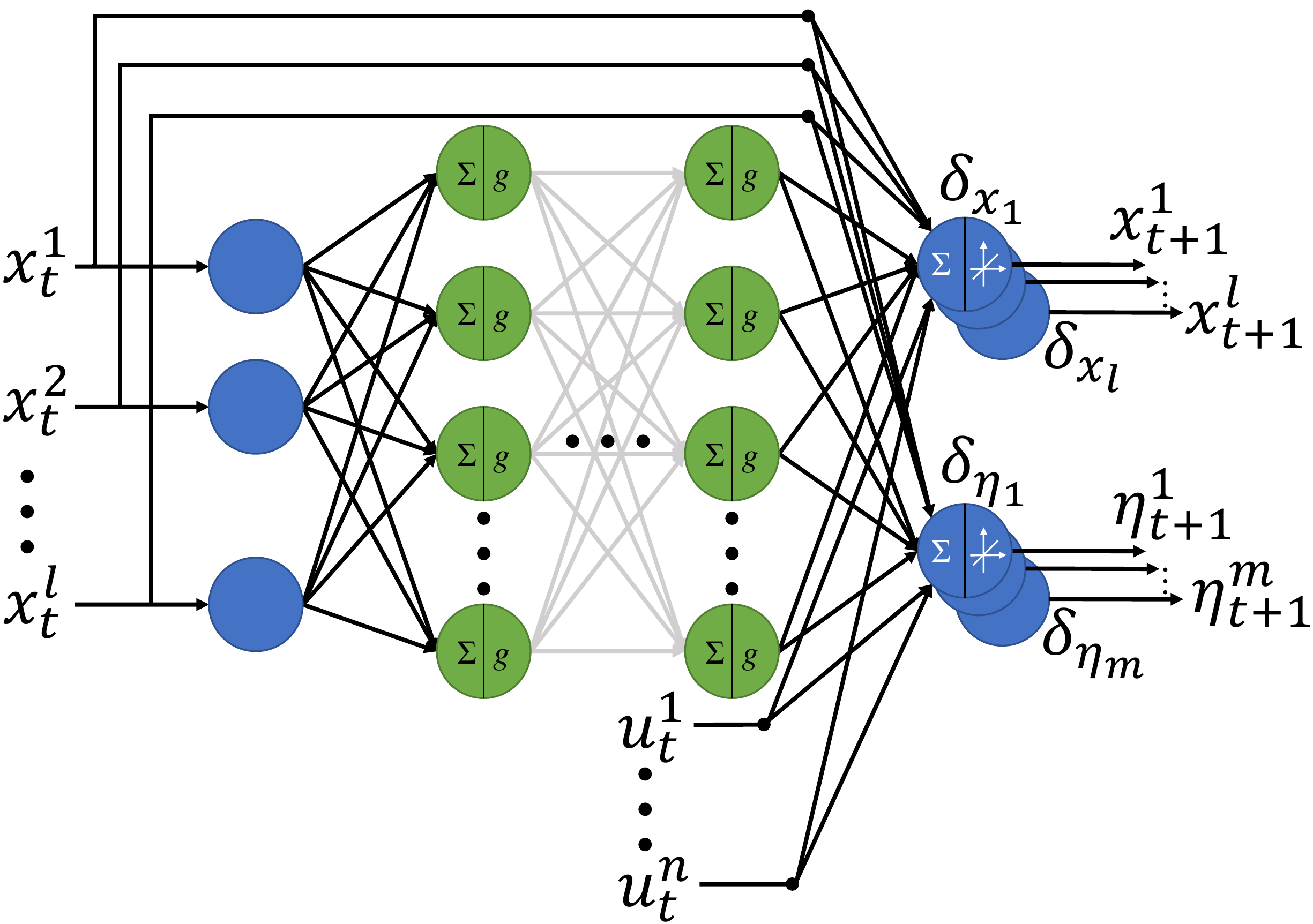}
    \caption{Diagram of the neural network and linear dynamic model to compute $x_{t+1}$ and $\eta_{t+1}$ given $x_t$ and $u_t$. With abuse of notation, we hereafter include $\zeta^*$ in $x$.}
    \label{fig:network}
\end{figure}

Consider the discrete-time dynamic system from (\ref{eq:disc-dyn-aut}). Let $g$ be a neural network, illustrated in Fig. \ref{fig:network}, defined by randomly initialized parameters $\theta$ to generate synthetic observables:
\begin{equation}
    \eta_t^\theta=g(x_t;\theta)
    \label{eq:eta-defn}
\end{equation}

\subsection{Extension to Anticausal Observables}
\label{sec:anticausal}

Define a datum vector $\xi_t^\theta \triangleq \left( x_t^\intercal, \eta_t^{\theta \intercal}, u_t^\intercal \right)^\intercal$. Let $A\in \mathbb{R}^{l\times p}$ and $H\in \mathbb{R}^{m\times p}$ be matrix coefficients modeling the state and augmented state transition dynamics, respectively. We override (\ref{eq:cases-dfl}) to include residuals in the original and augmented state dynamic equations:
\begin{equation}
    \begin{cases}
        x_{t+1}=A\xi_t^\theta+r_{x_{t+1}}^\theta \\
        \eta_{t+1}^\theta=H\xi_t^\theta+r_{\eta_{t+1}}^\theta \\
        \eta_0^\theta=g(x_0;\theta)
    \end{cases}
    \label{eq:cases-dfl-res}
\end{equation}

Given observation data of $x_t$, $x_{t-1}$, and $u_{t-1}$, we synthesize observations of the augmented state, $\eta_t^\theta=g(x_t;\theta)$ and $\eta_{t-1}^\theta=g(x_{t-1};\theta)$, and assemble datum vectors $\xi_{t-1}^\theta$.

By applying the discrete-time shift operator $T_{-1}$ to (\ref{eq:cases-dfl-res}) and rearranging, we can compute a residual for each observation: $r_{   x_t}^\theta=  x_t        -A\xi_{t-1}^\theta$ and $r_{\eta_t}^\theta=g(x_t;\theta)-H\xi_{t-1}^\theta$.

We define a quadratic loss function, $J_t(\theta,A,H)$, used to train the model:
\begin{equation}
    J_t(\theta,A,H)\triangleq r_t^{\theta \intercal} Qr_t^\theta
    \label{eq:cost}
\end{equation}
where $Q$ is a symmetric matrix coefficient and $r_t^\theta$ is a total residual given by
$r_t^\theta \triangleq \left( r_{x_t}^{\theta \intercal}, r_{\eta_t}^{\theta \intercal} \right)^\intercal$.

Model parameter matrices $A$ and $H$, as well as the parameters of the neural network, $\theta$, are computed by solving the following optimization problem via error backpropagation:
\begin{equation}
    \theta^o, A^o, H^o=\argmin_{\theta,A,H}\mathrm{E} \left[ J_t(\theta,A,H) \right]
    \label{eq:grad-desc}
\end{equation}

As discussed in \cite{asada-dfl}, augmenting the state with physical observables is often useful. Because this learned lifting linearization is data-driven, augmenting the state, $x$, with control-independent, physical observables is trivial. However, as reviewed in \cref{sec:bkgnd}, if the augmented state is dependent on the control input, the augmented system dynamics become anticausal. In state space modeling, output equations include inputs algebraically if there is a direct transmission term from inputs to outputs  \cite{kailath1980linear,derusso1997state}. Namely, observations of the system are functions of $x_t$ and $u_t$. Consider a vector of physical observables, $\zeta(x,u)\in \mathbb{R}^z$, suspected of including a dependence on $u$. As in \cite{korda2018linear,asada-dfl}, and \cite{mamakoukas2019local}, assume that this dependence is linear:
\begin{equation}
    \zeta(x,u)=\zeta^*(x)+Du
    \label{eq:zeta}
\end{equation}
where $\zeta^*(x)\in \mathbb{R}^z$ is exclusively a function of state and $D\in \mathbb{R}^{z\times n}$ is a fixed matrix coefficient of $u$. Assuming mean-zero data, because $\zeta^*(x)$ is uncorrelated with the control input, $\mathrm{E}\left[ \zeta^*(x) u^\intercal \right] =0$. Therefore, multiplying (\ref{eq:zeta}) by $u^\intercal$ and taking the expectation yields $\mathrm{E}\left[ \zeta(x,u)u^\intercal \right]=D\mathrm{E}\left[ uu^\intercal \right]$.

Given observations of $\zeta(x,u)$ and $u$, computing $D$ becomes a least-squares linear regression:
\begin{equation}
    \hat D=\mathrm{E}\left[ \zeta(x,u)u^\intercal \right]\mathrm{E}\left[ uu^\intercal \right]^{-1}
\end{equation}
assuming that the input is persistently exciting.

Before training the learned lifting linearization model in (\ref{eq:cases-dfl-res}), we preprocess the data to ``clean'' the physical observables from any linear dependence on $u$ via
\begin{equation}
    \zeta^*(x_t)=\zeta(x_t,u_t)-\hat Du_t
    \label{eq:clean-zeta}
\end{equation}
for each observation, $t$. Then, we augment the DFL model to include $\zeta^*(x)$. We override (\ref{eq:eta-defn}) with $\eta_t^\theta=g\left( \left( \zeta^{*\intercal}(x_t), x_t^\intercal \right)^\intercal; \theta \right)$ and follow the same training procedure described above to tune $g$, $A$, and $H$ using (\ref{eq:grad-desc}).

The complete learned lifted linearization algorithm is summarized in Algorithm \ref{alg}.
\begin{algorithm}
\SetAlgoLined
\KwResult{Lifting linearization of nonlinear dynamics}
 Randomly initialize neural network $g$ and linear dynamic model $A$, $H$\;
 $\hat D \gets \mathrm{E}\left[ \zeta u^\intercal \right]\mathrm{E}\left[ uu^\intercal \right]^{-1}$ \;
 \While{training}{
    get batch of $x_t$, $\zeta_t$, $u_t$, $x_{t+1}$, $\zeta_{t+1}$, $u_{t+1}$ from training dataset \;
    $\zeta^*_t \gets \zeta_t-\hat Du_t$ \;
    $\zeta^*_{t+1} \gets \zeta_{t+1}-\hat Du_{t+1}$ \;
    $\eta_t \gets g((x_t, \zeta^*_t))$ \;
    $\eta_{t+1} \gets g((x_{t+1}, \zeta^*_{t+1}))$ \;
    $\xi_t \gets \left( x_t^\intercal, \zeta^{*\intercal}_t, \eta_t^\intercal, u_t^\intercal \right)^\intercal$ \;
    $r \gets \left( ((x_{t+1}^\intercal, \zeta^{*\intercal}_{t+1})^\intercal-A\xi_t)^\intercal, (\eta_{t+1}-H\xi_t)^\intercal \right)^\intercal$ \;
    $J \gets r^\intercal Qr$ \;
    backpropagate $J$ to update $g$, $A$, and $H$ using Adam \;
 }
 $A_u \gets A_u+A_\zeta D$ \;
 $H_u \gets H_u+H_\zeta D$ \;
 \caption{Learned Lifting Linearization}
 \label{alg}
\end{algorithm}
\section{Numerical Examples}
\label{sec:toy}
\subsection{Toy Problem}

The modeling algorithm in \cref{sec:algo} is implemented in PyTorch \cite{pytorch} on a laptop running Ubuntu 18.04.5 LTS. The codebase is hosted as a git repository at \cite{repo}.

\begin{figure}[b]
    \centering
    \includegraphics[width=\linewidth]{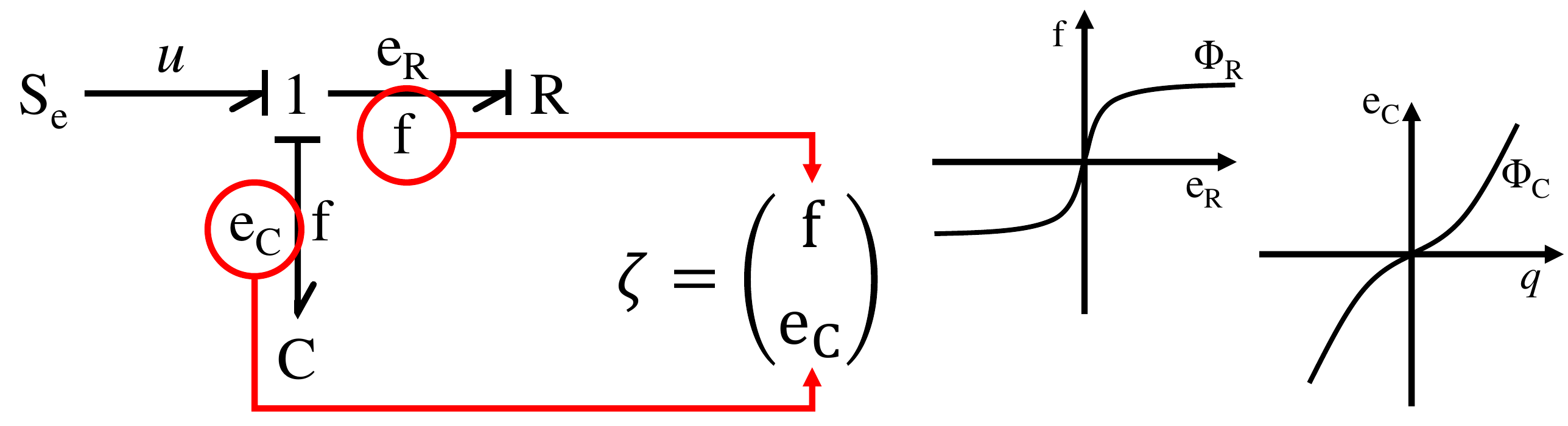}
    \caption{Bond graph of a nonlinear first-order system with state variable $q$.  In the bond graph, a nonlinear capacitor, C, and a nonlinear resistor, R, are connected to an effort source, $\mathrm{S_e}$, that is an exogenous input $u(t)$. Causality analysis of the bond graph determines that effort variable $\mathrm{e_C}$ is the output of the nonlinear capacitor, while the output of the nonlinear resistor is flow variable $f$. In the electrical circuit analogy, the effort variable $\mathrm{e_C}$ is the voltage across the capacitor, and the flow variable f is the current flowing through the resistor. They are connected with the exogenous input voltage $u(t)$ at the ``1'' junction, which is equivalent to Kirchhoff’s Voltage Law. The causality analysis also reveals that a direct transmission path exists from input $u(t)$ to flow variable f. Therefore, auxiliary variable f is not a causal variable for lifting the system. For more details about causality analysis, see \cite{asada-dfl,bond-graph-textbook}.}
    \label{fig:toy}
\end{figure}

\begin{figure}
    \centering
    \includegraphics[width=\linewidth]{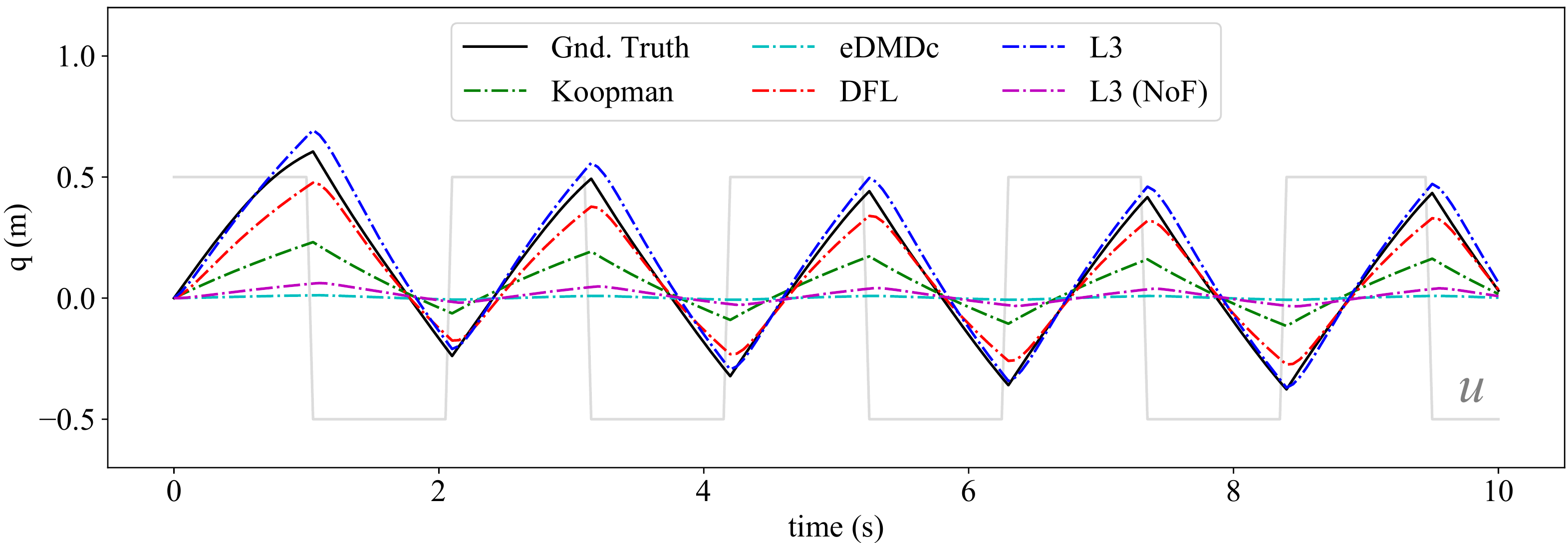}
    \caption{Results of open-loop simulation predicting the state, $q$ of the toy problem from Fig. \ref{fig:toy} excited by a square wave input (in gray). The solid black line indicates the true trajectory. The dashed lines indicate the open-loop simulated trajectories of various models: Koopman-with-control (Koopman, dim$\ =33$), extended dynamic mode decomposition with control (eDMDc, dim$\ =6$), dual-faceted linearization (DFL, dim$\ =4$), learned lifting linearization (L3, dim$\ =6$), and L3 without first filtering out the control input using (\ref{eq:clean-zeta}) (L3 NoF, dim$\ =6$).}
    \label{fig:results-toy}
\end{figure}

We test the learned lifting linearization (L3) algorithm on the nonlinear, massless spring-damper illustrated in Fig. \ref{fig:toy} with $\Phi_{\mathrm{R}}(\mathrm{e_R})=2/\left(1+e^{-4\mathrm{e_R}}\right)-1$, and $\Phi_\mathrm{C}(q)=\mathrm{sgn}(q)q^2$. We generate 100 5s trajectories at 20Hz with initial conditions and control inputs drawn from uniform random distributions. The state, $x$, consists only of the linear position, $q$, and the control input, $u$, is the scalar effort ($n=1$). We use the system bond graph to identify observables $\zeta=\left(\mathrm{f,e_C}\right)^\intercal$.

The neural network, $g$, approximating the optimal synthetic observables, $\eta$, is a fully connected network of 3 linear input neurons ($l=3$); two hidden layers, each with 256 ReLU neurons; and 2 output neurons, creating 2 synthetic observables ($m=2$). Before training, the data are randomly divided 80-20 into a training set and a validation set. The neural network, $g$, and the linear model consisting of $A$ and $H$ are trained in batches of 32 input-output pairs using an Adam optimizer \cite{adam} with $\alpha=10^{-5}$, $\beta_1=0.9$, $\beta_2=0.999$, and $\epsilon=10^{-8}$. The quadratic cost parameter $Q=I$. Before each training epoch, the learned lifting linearization model is evaluated using the validation dataset without backpropagating the loss. Training continues until the validation loss begins to increase.

\begin{table}[b]
    \centering
    \caption{Integrated squared error of the models used to simulate the toy problem over ten seconds.}
    \begin{tabular}{ccccc}
        Koopman & eDMDc & DFL & L3 & L3 (NoF) \\
        5.9 & 14 & 0.73 & 0.48 & 12
    \end{tabular}
    \label{tab:errors}
\end{table}

We benchmark the learned lifting linearization algorithm against Koopman-with-control, eDMDc, DFL, and L3 without the anticausal filter. Using the same data from the training and validation sets described above, 32 observables using polynomial basis functions are created for the Koopman observables and two similar observables are created for eDMDc. We train the eDMDc and DFL models using the same measurements, $\zeta$, as L3.

After training, we simulate all models given a zero initial condition and a square wave input trajectory. The modeled state trajectory is compared against the ground truth in Fig. \ref{fig:results-toy}. The learned lifting linearization model  outperforms the Koopman model despite the significantly lower dimensionality. The integrated squared errors of the simulated models are recorded in Table \ref{tab:errors}.

Note that the fidelity of the Koopman operator model is sensitive to hyperparameters. As discussed in \cite{soft-robot-koop}, without L1 regularization, high-dimensional Koopman models quickly overfit to the training data. Both DFL and L3 outperform eDMDc due to the anticausal filter compensating for the dependence of $\mathrm{e_R}$ on $u$. Without the anticausal filter, L3 performs only marginally better than eDMDc. L3 also has a slight advantage over DFL: in addition to penalizing nonlinearities in the state transition equation, the cost function of the learned lifting linearization in (\ref{eq:cost}) also penalizes nonlinearities in the augmented state transition equation.

% \begin{table}[b]
%     \centering
%     \begin{tabular}{|ccccc|}
%         1/0.8 & 0.06/0.03 & 0.04/0.2 & -0.01/0.1 & 0.001/0.001 \\
%         -0.8/0.8 & 1/2 & 1/0.2 & -0.4/-1 & 0.05/0.05 \\
%         & & & & \\
%         0.3/0.7 & -0.2/-0.1 & 0.5/0.2 & 0.1/-0.1 & -0.01/-0.01 \\
%         -0.7/-0.1 & 0.5/0.6 & 1/0.8 & 0.7/0.4 & 0.03/0.03
%     \end{tabular}
%     \caption{Matrices $A$ (top) and $H$ (bottom) of the linear dynamic model as computed for the toy problem using the LSE and EBP optimization methods, formatted [result from LSE]/[result from EBP].}
%     \label{tab:ldm}
% \end{table}

% We also use the toy problem to compare the least squares estimator and gradient-based approaches to learning $g$, $A$, and $H$. We implement both algorithms and compare the results in Table \ref{tab:ldm}. The results are numerically approximate, but simulations of the two methods reveal that the LSE approach results in far greater error. We hypothesize that the reason for the differing results comes from the inclusion of $r_{\eta_t}^\theta$ in the cost function in (\ref{eq:cost}). To test this hypothesis, we simulated both the LSE and EBP optimizations with $Q=\left( I, 0; 0, 0 \right)$ to remove the penalty on $r_{\eta_t}^\theta$. The resulting $A$ matrices were effectively equivalent. Importantly, these results indicate that the EBP optimization technique from \cref{sec:algo} is both less computationally expensive and more accurate than the LSE method.
\subsection{Excavation Process Modeling}
\label{sec:terra}

\begin{figure}
    \centering
    \includegraphics[width=\linewidth]{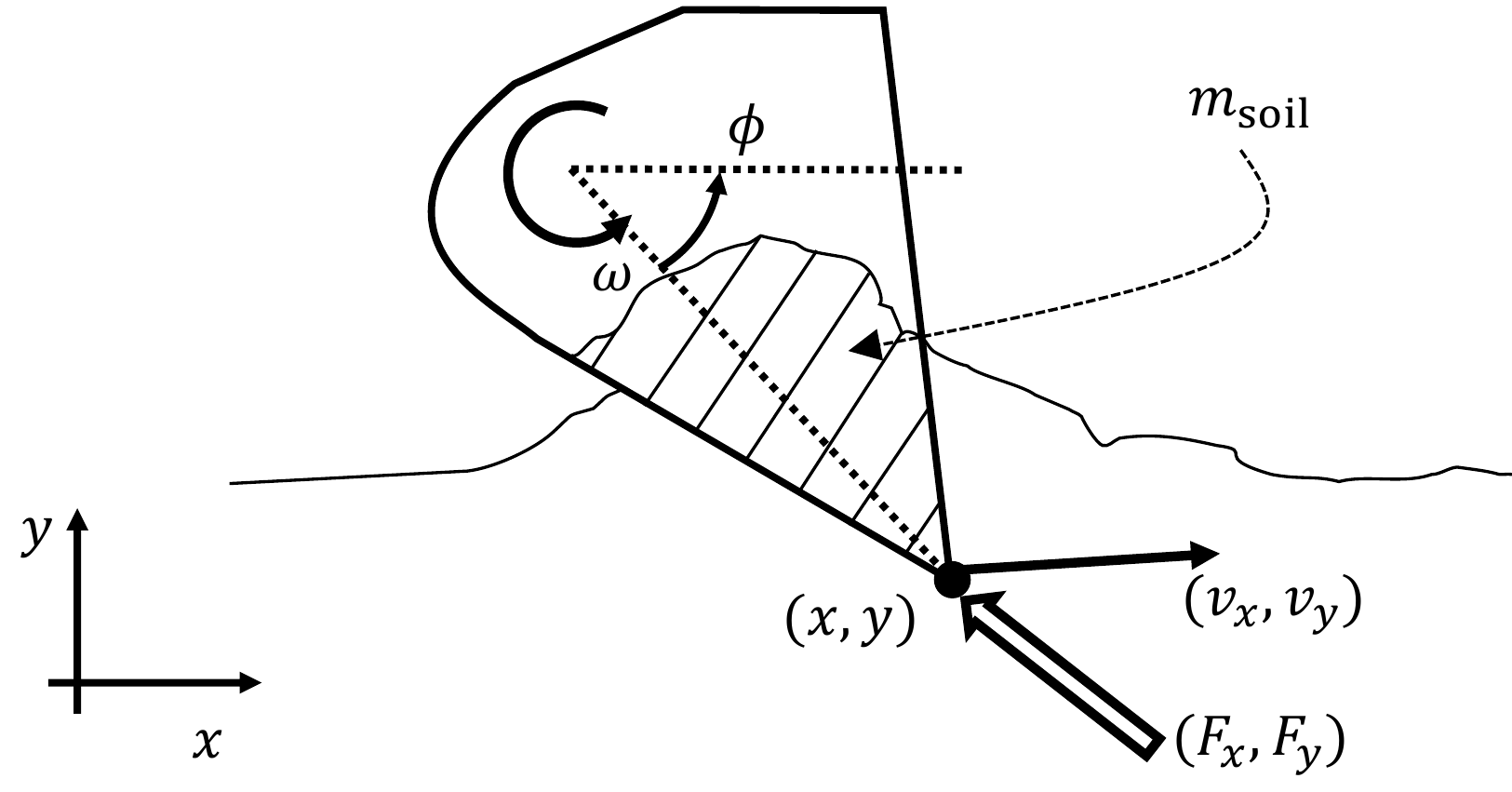}
    \caption{Diagram of the states and physical observables included in the data.}
    \label{fig:diagram-bucket}
\end{figure}

\begin{figure*}
    \centering
    \includegraphics[width=\linewidth]{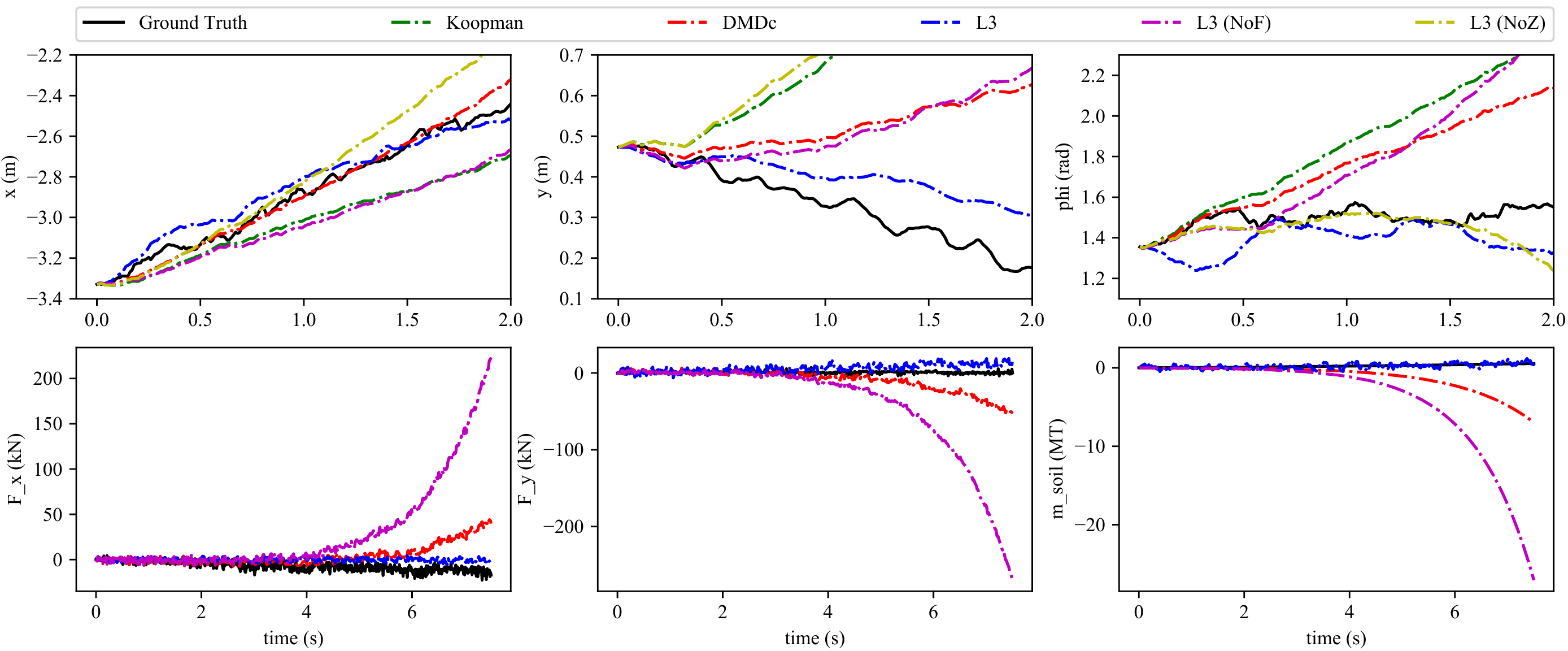}
    \caption{Results of open-loop simulation predicting the positional state and observable trajectories of the bucket through the soil given the initial condition and control input. The solid black line indicates the true, observed trajectory. The dashed green line indicates the trajectory predicted by the Koopman model. The dashed lines indicate the open-loop simulated trajectories of various models: dynamic mode decomposition with control (DMDc), learned lifting linearization (L3), L3 without first filtering out the control input using (\ref{eq:clean-zeta}) (L3 NoF), and L3 without any information from $\zeta$ (L3 NoZ).}
    \label{fig:results-dig}
\end{figure*}

We also test the learned lifting linearization algorithm on an autonomous excavator simulation using \textit{agxTerrain}, a specialized module of the AGX Dynamics \cite{agx} physics simulator used to test algorithms for autonomous excavation \cite{agx-1,agx-2,agx-3,agx-4} illustrated in Fig. \ref{fig:agx}. We generate a random soil profile by summing several 2-D Gaussians of random height and variance, yielding soil shapes like those in Fig. \ref{fig:agx}. Soil properties are set in accordance to the AGX ``gravel'' profile.

We collect 100 7.5s randomized trajectories from large sections of the workspace at 100Hz, setting one of them aside for testing. A diagram of the collected data is illustrated in Fig. \ref{fig:diagram-bucket}. The trajectories include six states, $x$: position along the x-axis, $\mathrm{x}$; position along the $y$-axis, $y$; bucket angle, $\phi$; velocity along the x-axis, $v_{\mathrm{x}}$; velocity along the $y$-axis, $v_y$; and rotational velocity of the bucket, $\omega$. The trajectories also include three control inputs, $u$: force along the x-axis, $u_{\mathrm{x}}$; force along the $y$-axis, $u_y$; and torque actuating the bucket, $u_\phi$. The trajectories also include three physical observables, $\zeta$: soil reaction force on the bucket along the x-axis, $F_{\mathrm{x}}$; soil reaction force on the bucket along the $y$-axis, $F_y$; and mass of the soil in the bucket, $m_{\mathrm{soil}}$.

These trajectories are generated using a naïve, noisy PID controller on the translation forces and bucket angle:
\begin{equation}
    \begin{aligned}
    u_{\mathrm{x}} & = \mathrm{PID}(\dot{\mathrm{x}} - \mathcal{U}(\dot{\mathrm{x}}_\mathrm{min},\dot{\mathrm{x}}_\mathrm{max})) + \mathcal{U}(-w_{\mathrm{x}},w_{\mathrm{x}}) \\ 
    u_y & = \mathrm{PID}(\dot y - \mathcal{U}(\dot y_{\mathrm{min}},\dot y_\mathrm{max})) + \mathcal{U}(-w_{y},w_{y}) \\ 
    u_\phi & = \mathrm{PID}(\phi - \mathcal{U}(\phi_{\mathrm{min}},\phi_{\mathrm{max}})) + \mathcal{U}(-w_{\phi},w_{\phi}) \\ 
    \end{aligned}
\end{equation}
where $\mathcal{U}$ is the uniform random distribution and $w_{\mathrm{x}}$, $w_y$, and $w_\phi$ are bounds on additional noise added to ensure persistent excitation. The set points are drawn from a uniform random distribution in accordance with \cite{Proctor2018}.

The learned lifting linearization and Koopman models for the terramechanics experiment are almost identical to those of the nonlinear spring-damper experiment, with some exceptions. The neural network, $g$, has 9 input neurons, 1 hidden layer with 256 ReLU neurons, and 4 output neurons. On average, training the L3 model took 2.5 hours. The domain of the Koopman dynamic model has a dimensionality of 67, compared to 16 for the learned lifting linearization model. We also benchmark against a DMDc model with a dimensionality of 12 trained using the observables, $\zeta$, in addition to the state. There was not a significant performance difference between DMDc and eDMDc. In addition to benchmarking the learned lifting linearization model against the Koopman and DMDc models, we also compare the results with and without filtering out the control input from the physical observables using (\ref{eq:clean-zeta}). In this model, instead of following the procedure described in \cref{sec:anticausal}, we incorporated $\zeta(x,u)$ directly into the state without filtering. Effectively, this forces the model to violate causality by predicting future values of observables dependent upon control input, $u$. Finally, we also test L3 without any information from observations, $\zeta$, to examine the effect of removing supplementary measurements.

After training the learned lifting linearization and Koopman models, we simulate both models using the control input trajectory and initial condition from the testing trajectory. The modeled trajectories of the positional states, x, $y$, and $\omega$, and the three observables, $F_x$, $F_y$, and $m_{\mathrm{soil}}$, are compared against the ground truth trajectories in Fig. \ref{fig:results-dig}. The integrated squared error across the three states for the learned lifting linearization model is only 5\% and 18\% of the same errors for the Koopman and DMDc models, respectively.

We also retrain the learned lifting linearization model \textit{without} first filtering out the control input using (\ref{eq:clean-zeta}) and simulate using the same technique, L3 (NoF). The integrated squared error across the three states is more than eight times the same error for the standard L3 model. Input filtering is vital to model robustness. A similar experiment performed without any information from supplemental measurements, L3 (NoZ), results in an error more than ten times greater than the standard L3 model. Various experiments retraining the L3 model without access to individual states or observables (left out of Fig. \ref{fig:results-dig}) result in similar reductions in accuracy.
% \addtolength{\textheight}{-0.5cm}
% This command serves to balance the column lengths on the last page of the document manually. It shortens the textheight of the last page by a suitable amount. This command does not take effect until the next page so it should come on the page before the last. Make sure that you do not shorten the textheight too much.
\section{Discussion and Conclusion}
\label{sec:conc}
% In this section, we highlight main takeaways from our experiments. 

In this paper, we presented a learned lifting linearization algorithm to model nonlinear dynamic systems. This model extended Koopman operator theory and dual-faceted linearization by training a neural network to produce nonlinear observables to augment the state. We also presented an algorithm to ``clean'' anticausal physical observables of any linear dependence on control input so that they can be used by the neural network to generate richer synthetic observables. We tested this algorithm on a nonlinear, massless spring-damper model and an autonomous excavation simulation, and we compared the results against Koopman, DMD, and DFL models. Learned lifting linearization outperformed all benchmarks at minimizing state prediction error.

As with many data-driven techniques, training data quality is paramount to model accuracy. The reduced performance in modeling progression along the $y$-axis is likely due to the small domain of the training data along that dimension. If we initialize the system in a different configuration from what was recorded during data collection, or if the quantity of data available for learning is reduced by more than half, models such as L3 and DMD perform substantially worse.

In real-world excavation tasks, soil properties can vary. While this paper addresses more homogeneous soil profiles like gravel, intelligent use of observables to include information about the environment could enable L3 to learn optimal auxiliary variables and linear models for dynamic soil properties. Additionally, L3 could be primed with simulator data to reduce the amount of hardware-in-the-loop learning required.

Excavators often require expensive hydraulics modifications to execute end effector force control. State-of-the-art excavators have solved this problem \cite{jud2019autonomous}. This work focused on the nonlinearities involved in  soil dynamics. Future work on this topic should model the nonlinearity of hydraulic systems.

\bibliographystyle{ieeetr}
\bibliography{main}


% \end{IEEEbiography}


% % insert where needed to balance the two columns on the last page with
% %\newpage



%\vfill


\end{document}